# GCFX: Generative Counterfactual Explanations for Deep Graph Models at the Model Level

**Jinlong Hu \*, Jiacheng Liu**

Guangdong Provincial Key Laboratory of Multimodal Big Data Intelligent Analysis

School of Computer Science and Engineering

South China University of Technology, Guangzhou, China

jlhu@scut.edu.cn

**Abstract**:

Deep graph learning models have demonstrated remarkable capabilities in processing graph-structured data and have been widely applied across various fields. However, their complex internal architectures and lack of transparency make it difficult to explain their decisions, resulting in opaque models that users find hard to understand and trust. In this paper, we explore model-level explanation techniques for deep graph learning models, aiming to provide users with a comprehensive understanding of the models' overall decision-making processes and underlying mechanisms. Specifically, we address the problem of counterfactual explanations for deep graph learning models by introducing a generative model-level counterfactual explanation approach called GCFX, which is based on deep graph generation. This approach generates a set of high-quality counterfactual explanations that reflect the model's global predictive behavior by leveraging an enhanced deep graph generation framework and a global summarization algorithm. GCFX features an architecture that combines dual encoders, structure-aware taggers, and Message Passing Neural Network decoders, enabling it to accurately learn the true latent distribution of input data and generate high-quality, closely related counterfactual examples. Subsequently, a global counterfactual summarization algorithm selects the most representative and comprehensive explanations from numerous candidate counterfactuals, providing broad insights into the model's global predictive patterns. Experiments on a synthetic dataset and several real-world datasets demonstrate that GCFX outperforms existing methods in terms of counterfactual validity and coverage while maintaining low explanation costs, thereby offering crucial support for enhancing the practicality and trustworthiness of global counterfactual explanations.

## 1. Introduction

Deep graph learning (DGL) aims to apply deep learning techniques to automatically learn the topology and attribute information of complex graph data. This process generates expressive embedded representations that effectively enhance the performance of downstream analytical tasks [1–3]. Currently, the message passing mechanism has become a widely adopted paradigm in deep graph learning [4,5]. A series of Graph Neural Networks (GNNs) derived from this mechanism have achieved significant breakthroughs in graph characterization and analysis tasks, demonstrating excellent performance in areas such as drug molecule discovery, biomedical diagnosis, financial transaction analysis, and recommender systems [1,6–8]. However, GNNs, like other deep neural network models, function as black boxes for users, who typically receive only predictive results for the target task without insight into the decision-making processes embedded within the numerous network parameters [9–11]. The opacity of these black-box models can lead to unfair or incorrect decisions due to potential biases and sensitive information, undermining user trust and hindering their adoption in critical applications involving fairness, privacy, and security. Therefore, it is essential to provide human-understandable explanations alongside accurate predictions to ensure that users can deploy deep models with confidence and reliability [12–14].

Techniques for model explainability generally investigate the fundamental relationships that influence a model's predictions, elucidate the underlying mechanisms of its decision-making process, evaluate the conditions and extent to which the model relies on these relationships, and detect potential biases or prejudices. These methods ultimately enhance the reliability and credibility of the model's predictions [15–17]. In recent years, with the rapid advancement of explainability approaches in the broader field of deep learning, there has been growing scholarly interest in the explainability of deep graph learning models. Interpretability techniques for deep graph models can be categorized based on the scope of explanation into local, instance-level explanations, which provide attribution analyses for the predictions of individual samples, and global, model-level explanations, which are independent of specific samples and aim to uncover overarching, generalizable insights into the model's predictive behavior [10,18,19]. Furthermore, depending on the intended purpose of the explanation, these methods can be divided into factual explanations, which identify the sample features or substructures most pertinent to the prediction [20], and counterfactual explanations, which aim to delineate the model's decision boundaries by introducing minimal perturbations to samples to change their predicted outcomes [21].

Counterfactual analysis addresses the limitations of factual attribution by explaining how changes in input lead to different model predictions. It clarifies why a model produces certain outputs by highlighting which input modifications result in altered predictions, helping practitioners identify and correct potential biases in models and thereby improving fairness and reliability in decision-making [22, 23]. Recent research on counterfactual explanatory methods primarily focuses on instance-level explanations, which depend heavily on specific input samples and provide localized counterfactual examples. However, these instance-level explanations often fail to offer generalized insights into the overall decision-making logic of black-box models. To overcome this limitation, model-level counterfactual explanations aim to identify a small set of representative counterfactuals that can explain the global behavior of the model. Recent approaches rely on heuristic random search techniques to find a limited number of counterfactuals that cover as many input samples as possible [30]. Nevertheless, explanations generated by these methods frequently struggle to satisfy feasibility constraints in real-world domains and lack robust generalization performance on unseen data. Additionally, the large number of model queries required by these methods raises concerns regarding cost and efficiency, which is particularly critical for domain-specific black-box models with expensive access.

In this paper, we propose a model-level counterfactual explanation method based on deep graph generation, called the Global Counterfactual Explainer (GCFX). This method employs a generative network architecture to learn the potential distributions of both the input real data and its counterfactuals. Additionally, it incorporates a multi-objective loss function and a global counterfactual summarization algorithm to provide black-box models with a set of high-quality, model-level counterfactual explanations that capture the global predictive behavior.

## 2. Related Works

Recently, explainable artificial intelligence (XAI) has generally categorized the explainability of deep models into two types: pre-existing model self-interpretation (ante-hoc) and post-hoc model explainability [18, 24]. Self-interpretability typically refers to the model's architecture design or learning process itself providing a certain degree of interpretability, allowing the model's decision outputs and behaviors to be directly understood. Examples include decision trees, logistic regression, and attention models. In contrast, post-hoc explainability involves using external tools or techniques to explain the decision-making process and behavior of the model after training is complete. These methods are often model-agnostic, meaning they do not rely on the internal structure of the model or access to its parameters or weights, enabling users to reliably explain the behavior of any black-box model. Furthermore, depending on the level of explanation, methods for explaining deep graph models can be divided into local instance-level explanations, which provide attribution analysis for specific sample predictions, and global model-level explanations, which do not depend on specific samples but aim to reveal the model's overall predictive behavior and generalized insights [18, 19]. Based on the explanation's objective, explainability methods can also be categorized into factual explanations, which identify the sample features or substructures most relevant to the prediction, and counterfactual explanations, which seek to reveal the model's decision boundaries by minimally perturbing samples to change their predictions [21].

XGNN [19] represents the first effort in model-level explanation for deep graph learning, providing factual explanations by generating key subgraph patterns. It trains a graph generator using reinforcement learning (RL) with policy gradient optimization, combining feedback from the graph model with a reward function designed based on artificial graph rules. DGX [20] can produce diverse explanations by generating a set of distinguishable graphs and offers customized explanations based on prior knowledge or user-specified constraints. GNNInterpreter [25] is a generative global fact interpreter that employs a numerical optimization method to generate model-level explanatory graphs, maximizing the likelihood that these graphs are predicted by the model to belong to the target category. Additionally, it introduces a cosine distance constraint based on graph embeddings to prevent the explanation from deviating from the true data distribution and incorporates the Gumbel-Softmax reparameterization technique to enable gradient backpropagation during optimization for discrete data. GCExplainer [26] is inspired by the concept-based image explanation method ACE [27], which identifies abstract concepts by clustering the node embeddings from the last layer of the GNN output and provides a global explanation. Each concept and its importance are represented by different clusters and the number of nodes contained within them, respectively. GLGEexplainer [28] aggregates multiple local explanations from PGExplainer [29] to derive a global explanation by projecting local explanation maps onto a set of learnable concepts and training an entropy-based logically explainable network. This network ultimately outputs conceptual logical formulas for each category as a summary of the global explanation.

In deep learning explainability research, counterfactual analysis has recently garnered significant interest. Unlike factual explanation techniques, which focus on interpreting a model's decisions based on input features, counterfactual methods investigate how models cross decision boundaries. These methods aim to reveal the logic and sensitivities underlying model behavior by generating hypothetical scenarios that differ slightly from actual cases. Consequently, counterfactual explanations provide insights into model decision boundaries and inherent biases by identifying critical differences in the sample space that can alter model predictions. Currently, GCFExplainer [30] is the only approach addressing model-level counterfactual explanations. It explains all input samples associated with a particular predicted behavior, structures the counterfactual space near these inputs as a meta-graph, employs graph editing operations to perform vertex-enhanced random walks [31] on this meta-graph to identify candidate counterfactuals, and then selects a small subset with maximal coverage from these candidates as the global explanation. Although counterfactual methods have significantly enhanced model interpretability, generating accurate, high-quality counterfactual examples remains a challenging task. This challenge stems from the need to thoroughly understand both the underlying input data distribution and the model's internal mechanisms.

## 3. Method
### 3.1 Overview of Generative Model-Level Counterfactual Explanation Methods

Given a labeled graph dataset $\mathcal{D} = \{(\mathcal{G}_i, Y_i) | 1 \leq i \leq D\}$, and a deep graph classification model $\mathcal{M}: \mathcal{G} \to Y$ trained on this dataset, for any input sample $\mathcal{G} = (A, X)$ in the dataset, let its true label be $Y_t$, and the model's predicted output label be $\mathcal{M}(\mathcal{G}) = Y_p$. The counterfactual explanatory method aims to find counterfactual examples $\mathcal{G}^* = (A^*, X^*)$ that are as similar as possible to sample $\mathcal{G}$, while satisfying $\mathcal{M}(\mathcal{G}^*) \neq Y_p$.

Given a cost threshold $\delta$, we define the approximate counterfactual of the model's prediction sample $\mathcal{G}$ as the set of all counterfactual examples that satisfy Eq. (1), where the function $\omega(\cdot)$ measures the cost of perturbation from input sample to its counterfactual examples.

$$CC(\mathcal{G}) = \{\mathcal{G}^* \mid \mathcal{M}(\mathcal{G}^*) \neq \mathcal{M}(\mathcal{G}) \wedge \omega(\mathcal{G}^*, \mathcal{G}) \leq \delta\} \tag{1}$$

For the target dataset label $Y_t$ and its subset of samples $\mathcal{D}^{Y_t} = \{\mathcal{G}_i \mid \mathcal{M}(\mathcal{G}_i) = Y_t\}$, by generating a sample approximate counterfactual for each sample, we obtain a candidate set of approximate counterfactuas for the model's prediction of the target label $\Psi(Y_t)$.

$$\Psi(Y_t) = \bigcup_{\mathcal{G}_i \in \mathcal{D}^{Y_t}} CC(\mathcal{G}_i) \tag{2}$$

Model-level counterfactual explanation aims to use as few approximate counterfactuas as possible to explain as many target samples as possible. Thus, the model-level counterfactual explanation of the model's predicted label $Y_t$ can be described as a small subset of both "representative" and "diverse" counterfactuals selected from the candidate set of approximate counterfactuas $\Omega \in \Psi(Y_t)$. Here, the term "representative" means that each global counterfactual is general enough to explain as many input samples as possible, while "diversity" means that each global counterfactual is unique enough to cover as many different input samples as possible. To this end, this chapter adopts the following model-level counterfactual optimization objective: to find the smallest possible subset of approximate counterfactual candidates $\Omega = \{\mathcal{G}_1^*, \cdots, \mathcal{G}_K^*\}$ that maximizes the explanatory coverage of the target sample set $\mathcal{D}^{Y_t}$. The optimization objective is shown in Eq. (3), where the function $Coverage_\delta(\cdot)$ evaluates the explanatory coverage of the target sample set by a subset of counterfactuals within the cost threshold δ. This objective naturally encourages the explanation algorithm to eliminate redundancy among the counterfactual candidates for a particular instance, resulting in global explanations with higher information density.

$$\arg\max_{\Omega} \boldsymbol{Coverage}_\delta(\Omega, \mathcal{D}^{Y_t}) \quad s.t. \quad |\Omega| \ll |\mathcal{D}^{Y_t}| \qquad (3)$$

According to the optimization objective of model-level counterfactuals, we propose a generative model-level counterfactual explainer (GCFX), which consists of two phases: local approximate counterfactual explanatory graph generation and global counterfactual summarization. First, given the trained deep graph model $\mathcal{M}(\cdot)$ and the target label $Y_t$, the counterfactual graph generation model is constructed and trained on the target sample set $\mathcal{D}^{Y_t}$. Specifically, a counterfactual graph generation model based on the variational auto-encoder architecture (VQ-CFX) is developed. VQ-CFX introduces vector quantization to implement a local structural tagger and combines a dual graph encoder with a message passing neural network (MPNN) [56] decoder to generate approximate counterfactual examples. Second, VQ-CFX is applied to generate approximate counterfactual examples for each target sample, producing the local counterfactual candidate set for the target label, denoted as $\Psi(Y_t)$. Then, a global counterfactual summarization algorithm (GCFS) is used to filter the best model-level counterfactual explanations from the candidate set Ω. The GCFS algorithm aims to maximize global coverage by selecting an optimal subset of model-level counterfactual explanations.

## 3.2 Approximate Counterfactual Graph Generation

To generate approximate counterfactual graphs for a target sample set, we develop a deep graph generation network based on an enhanced variational graph auto-encoder (VGAE). The traditional VGAE is an encoder-decoder neural network architecture combined with variational inference, where the encoder maps each input graph to a latent variable space and approximates it with a prior distribution. The decoder then samples latent representations from this prior distribution to generate graph samples. Traditional VGAEs learn continuous posterior distributions from the data and enforce their approximation to a normal prior distribution, which leads to issues such as information redundancy and posterior collapse. These problems impair the model's ability to learn meaningful latent representations, resulting in generated samples that lack diversity. Furthermore, traditional VGAEs typically encode and decode based on graph-level embedded representations, making it challenging for decoders to accurately capture and reconstruct fine-grained node features and structural information.

To address the aforementioned challenges, we integrate the vector quantization technique to enhance VGAE and propose a counterfactual explanatory graph generation method, VQ-CFX, as illustrated in Figure 1. First, two graph encoders are employed to learn the factual and counterfactual latent representations of the input graph, respectively. Next, the vector quantization technique [55] is applied to discretely encode the continuous latent vector space, producing discrete quantized representations of the input graph. Finally, a MPNN decoder generates the corresponding target graph samples based on these quantized representations. The VQ-CFX model simultaneously reconstructs input samples and generates counterfactual examples. It is trained using a specially designed multi-objective loss function to produce approximate counterfactual interpretation graphs that align with the true data distribution. We provide a detailed description of the VQ-CFX model's architecture and underlying principles by focusing on three key components: the graph encoder, vector quantization, and MPNN decoder.

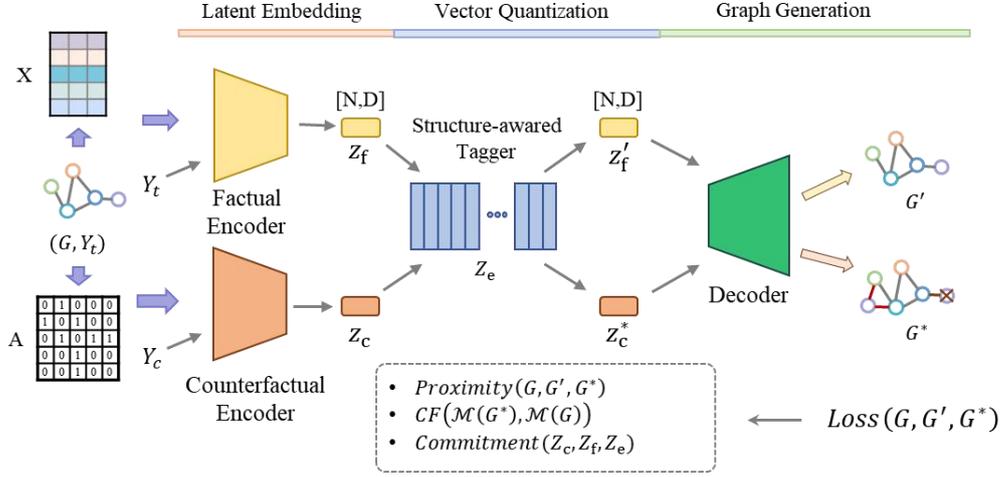

Figure 1 The architecture of generative model VQ-CFX

### 3.2.1 Graph Encoder

The VQ-CFX model comprises two graph encoders: the factual encoder and the counterfactual encoder. These encoders are designed to learn the true posterior distribution and the counterfactual distribution of the input data, respectively. Both graph encoders share a similar architecture, constructed by stacking multiple building blocks. Each block consists of a graph neural network layer, a normalization layer, and an activation function. Specifically, the graph neural network layer employs the Graph Isomorphism Network (GIN); the normalization layer uses GraphNorm [56]; and the activation function is uniformly the Rectified Linear Unit (ReLU). Additionally, the two graph encoders in VQ-CFX share a Label Encoder (LE), which incorporates sample labels as conditional prior distributions to facilitate potential representation learning. The architecture of the encoder network is illustrated in Figure 2.

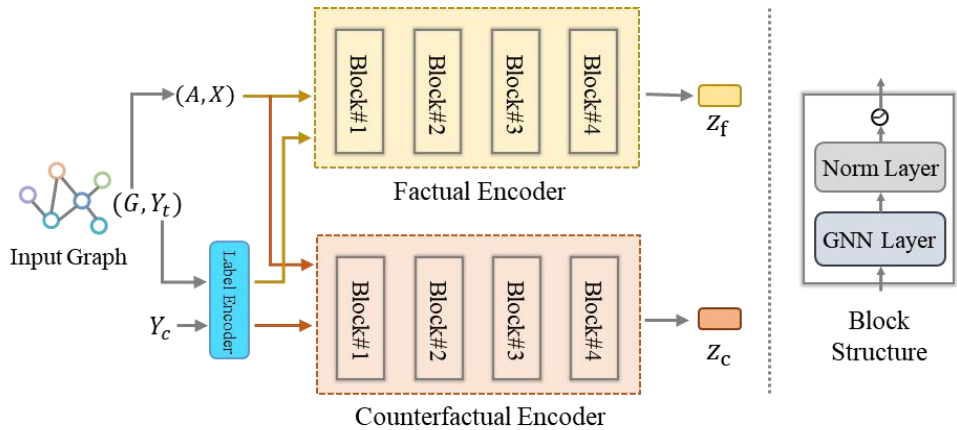

Figure 2　The architecture of dual Encoder network and Block

For the input sample $\mathcal{G} = (A, X)$ with target label $Y_t$, assume it has N nodes and F-dimensional continuous node features (discrete features are transformed into continuous features using one-hot encoding). The output dimension of the label encoder is $D_{LE}$ and the output dimension of the graph neural network layer is $D_{Lat}$. First, the label encoder maps each discrete label into a continuous conditional vector $v$, which is concatenated with the node feature matrix $X$ by rows and input into the first building block of the graph encoder, as shown in Eq. (4).

$$\begin{aligned} v &= LE(Y) \\ H^{(0)} &= [X; v] \end{aligned} \qquad (4)$$

Here, $v \in \mathbb{R}^{D_{LE}}$, $X \in \mathbb{R}^{N \times F}$, $H^{(0)} \in \mathbb{R}^{N \times (F+D_{LE})}$, The symbol [ ; ] denotes the concatenation operator. In each block layer, the graph neural network continuously updates node representations through message passing and neighborhood aggregation mechanisms. The normalization layer then normalizes the features of all node representations within the samples. Finally, the node embeddings are obtained after applying the activation function. For the input adjacency matrix $A$ and feature matrix $H^{(0)}$, the node representation update process at the $l$-th layer of the GIN is shown in Eq. (5).

$$H^{(l)} = (A + (1 + \epsilon) \cdot I) \cdot H^{(l-1)} \cdot W^{(l)} \qquad (5)$$

The inputs of the two graph encoders in the VQ-CFX model share the same adjacency matrix and node features, with the difference being the reception of different labels as condition variables to learn the potential representations of the samples under different conditional distributions. In particular, the fact encoder receives the sample true label $Y_t$, and the counterfactual encoder receives the counterfactual label $Y_c$ ($Y_c \neq Y_t$), which outputs the factual potential representation $Z_f$ and the counterfactual potential representation $Z_c$, respectively, after the learning update of the multilayer Block, which further allows the model decoder to generate the counterfactual example of a true sample along with reconstructing it.

The two graph encoders in the VQ-CFX model share the same adjacency matrix and node features as inputs. The key difference lies in the conditional labels they receive, which enables the model to learn potential representations of samples under different conditional distributions. Specifically, the factual encoder receives the true sample label $Y_t$, while the counterfactual encoder receives the counterfactual label $Y_c$ (where $Y_c \neq Y_t$). After processing through the multilayer block, these encoders output the factual potential representation $Z_f$ and the counterfactual potential representation $Z_c$, respectively. This design allows the model decoder to generate a counterfactual example corresponding to a true sample, as well as reconstruct the original sample.

### 3.2.2 vector quantization

In the Vector Quantization (VQ) component, the VQ technique aims to map a continuous vector space into a finite discrete set, enabling effective compression and efficient encoding of the latent data distribution. In generative models, vector quantization partitions the continuous latent distribution learned by the model into a finite number of discrete regions, which are organized into a fixed-size codebook. Each region corresponds to a vector in the codebook, called a codevector, representing distinct high-level local features. For successive latent representations of arbitrary inputs, the vector quantization step involves nearest neighbor matching to find the closest codeword in the codebook and replacing the original latent vector with that codeword. Additionally, vector quantization employs the straight-through estimator to address the non-differentiability of this replacement process, allowing both the generative model and the codebook to be updated via backpropagation.

The VQ-CFX eliminates both the pooling operation in the graph encoder and the variational inference process. Instead of modeling latent data distributions based on the dimensions of sample graph embeddings, as traditional variational autoencoders do, it focuses on learning local structural patterns with diverse high-level semantics directly from sample node-level latent representations. These discrete local latent representation vectors are then quantized and stored using a global, structure-aware tagger. This approach addresses the limitations of traditional variational autoencoders, which model potential data distributions at the graph level. In traditional models, the learned posterior distributions heavily depend on the representational and decoding capabilities of the encoder and decoder, often leading to posterior collapse when their performances are mismatched. Furthermore, approximating the posterior distribution to an assumed Gaussian prior through Bayesian inference may introduce an excessively strong inductive bias, preventing the model from adequately learning and representing the true data distributions.

Specifically, VQ-CFX introduces a codebook $\mathcal{C}$ with a specified size $N_{vq}$ and vector dimension $D_{Lat}$. The primary objective of this codebook is to learn an expressive local structural representation space with efficient compression. This space can be indexed by different codewords to capture the diverse local structural patterns within the node neighborhood. Taking the counterfactual potential representation $Z_c \in \mathbb{R}^{N \times D_{Lat}}$ as an example, the quantization process proceeds as follows: for each row of the potential vector $z$ in $Z_c$, select the closest codevector $z_q$ from the codebook using the nearest-neighbor matching strategy described in Eq. (6); replace the original vector $z_q$ with $z$, and preserve the gradient information in $z$ by applying the Straight-through technique as outlined in Eq. (7). After completing all substitutions, the resulting output is the counterfactual quantized representation of the sample $Z_c^*$.

$$z_q = \underset{e \in \mathcal{C}}{\arg\min} \|z - e\| \tag{6}$$

$$z^* = z + sg[\![z_q]\!] - sg[\![z]\!] \tag{7}$$

Where $\mathcal{C}$ is the codebook, $e$ is the codevector in the codebook, and the function $\|\cdot\|$ denotes the distance between the two vectors. $z^*$ is the vector obtained after the substitution operation, and the function $sg[\![\cdot]\!]$ denotes the stopping of the gradient operation.

It is essential to recognize that, in conventional graph generation tasks, the codebook primarily captures the latent distribution of local structures within the input data. In contrast, for the global counterfactual generation task examined in this study, VQ-CFX employs the codebook to simultaneously learn all local structural patterns present in both the true data distribution and its corresponding counterfactual distributions. This is accomplished by performing input sample reconstruction and counterfactual example generation concurrently during the model's training phase.

### 3.2.3 MPNN Decoder

We propose a message passing neural network (MPNN)-based decoder architecture to replace the linear decoder (e.g., MLP) in the traditional variational graph self-encoder model. The MPNN [56] is a graph neural network framework that learns node and graph representations by passing messages and aggregating information over graph-structured data. The MPNN decoder iteratively refines the adjacency matrices and node features using a multilayer message-passing architecture to generate target graph samples, as illustrated in Figures 3. Specifically, based on the quantized node-level latent representations and conditional variables, the decoder first initializes the reconstructed node features and adjacency matrices. Then, at each decoding step, it alternately updates the reconstructed node and adjacency matrix representations using the node reconstruction module and the edge reconstruction module. Finally, it outputs discretized adjacency matrices and reconstructed node feature matrices through two separate predictors.

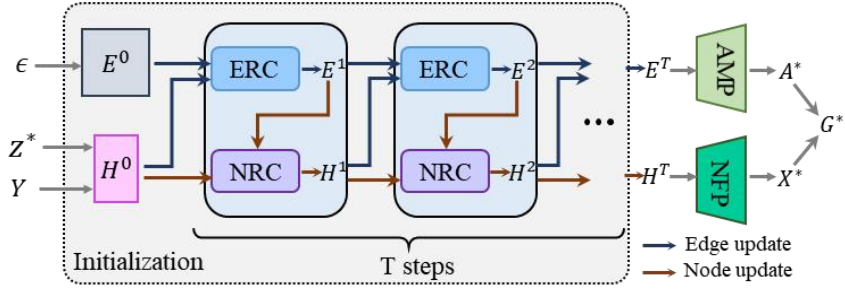

Figure 3 The architecture of MPNN decoder network

The MPNN decoder is designed to generate high-quality graph samples through a joint iterative optimization process within the message-passing network architecture. This approach enables the VQ-CFX model to perform both real sample reconstruction and counterfactual example generation, leveraging two potential representations: the inputs and the condition variables. Specifically, given the input counterfactual potential representation $z^*$ and label $Y_t$, the decoder first initializes the node feature reconstruction representation $H^0 \in \mathbb{R}^{N \times D_0^n}$ and randomly samples the edge reconstruction representation from a specified noise distribution $\epsilon \sim N(0,1)$. Subsequently, higher-order correlations between nodes and edges are progressively learned and integrated through the $T$ layer decoding network. The outputs $E^T$ and $H^T$ parameterize the original distributions of the adjacency matrices and node features of the target samples, respectively. Finally, the adjacency matrix predictor and node feature predictor produce the decoded adjacency matrix $A^*$ and node features $X^*$, yielding the final generated graph $G^* = (A^*, X^*)$.

Each layer of the decoding network consists of an edge reconstruction module (ERC), which treats $E^0$ as a fully connected neighbor matrix representation and updates the reconstructed representation of each potential edge using a two-layer MLP network, and a node reconstruction module (NRC), which updates the potential feature representation of the current node using a message-passing neural network designed to aggregate all neighboring node features and their corresponding edge features. For the $t$-th layer of the decoding network $(1 \leq t \leq T)$, the update process of its ERC and NRC modules is expressed in Eq. (8).

$$\begin{aligned} E_{ij}^t &= ERC([H_i^{t-1}; E_{ij}^{t-1}; H_j^{t-1}]) \\ H_j^t &= NRC\left(H_j^{t-1}, \sigma_{i \in \mathcal{N}_j}(H_i^{t-1}, E_{ij}^t)\right) \end{aligned} \tag{8}$$

where $E_{ij}$ denotes the edge reconstruction representation of the potential connection between node $i$ and node $j$, $H_i$ denotes the feature reconstruction representation of node $i$, and the symbol $[\ ]$ denotes the vector connection operation. $\mathcal{N}_j$ denotes the set of neighboring nodes of node $j$. Here, all other nodes in the sample are considered, and the function $\sigma(\cdot)$ is used to transmit and aggregate the neighboring messages of the current node.

Finally, both the adjacency matrix predictor (AMP) and the node feature predictor (NFP) are implemented using a single-layer fully connected network. Specifically, the output dimension of the NFP matches the dimension of the node features in the input samples. If the input samples include $m$ types of node connections (edge types), an additional type is defined to indicate the absence of any connection between nodes. The output dimension of the AMP is $m+1$.

### 3.2.4 Objective function and training

The VQ-CFX model is trained by minimizing an objective function $\mathcal{L}$ comprising three loss terms: proximity loss, counterfactual loss, and vector quantization loss. This objective function is calculated using Equation (9) as a weighted sum of the individual loss components.

$$\mathcal{L} = \alpha_1 \mathcal{L}_{px} + \alpha_2 \mathcal{L}_{cf} + \alpha_3 \mathcal{L}_{vq} \tag{9}$$

First, the proximity loss trains the two encoders and decoders to generate graphs that approximate both the input samples and the target counterfactuals. It includes losses for reconstructing the input samples from the factual latent representations $Z'_f$, as well as for generating the approximate counterfactual examples from the counterfactual latent representations $Z^*_c$. For the latter, we subtract the relaxation factor $\theta$ and replace the real input sample with $\mathcal{G}'$ in order to allow for a small variance in the counterfactual example. For the sample $\mathcal{G}^m = (X^m, A^m)$ and $\mathcal{G}^n = (X^n, A^n)$, we combine the node reconstruction error and topology reconstruction error to compute the proximity loss, as shown in Eq. (10). Here, N is the number of nodes in the input graph instance, $\gamma \geq 1$ is the scaling factor of the cosine error of the node features, the function $\psi(\cdot)$ evaluates the cosine similarity between the corresponding node features, and the function $\phi(\cdot)$ serves to further impose a penalty on the larger value of the absolute errors of the adjacency matrix.

$$\mathcal{L}_{px}(\mathcal{G}^m, \mathcal{G}^n) = \frac{1}{N} \sum_i^N (1 - \psi(X_i^m, X_i^n))^\gamma + \phi(|A^m - A^n|) \tag{10}$$

$$\phi(x) = \|x + 0.5\|_2^2 \tag{11}$$

Counterfactual loss is used to guide the VQ-CFX to perturb the input samples with counterfactuals, which makes the predictor make different predictions from the input samples. The counterfactual loss is implemented by binary cross entropy function calculation and counterfactual labeling as in Equation (12).

$$\mathcal{L}_{cf} = CrossEntropy(\mathcal{M}(\mathcal{G}^*), Y_c) \tag{12}$$

The vector quantization loss is primarily used to train the encoder and update the codebook. Its objective is to ensure that each individual code vector in the codebook provides a better high-level local representation. Simultaneously, it encourages the latent representation output by the encoder to be close to the corresponding code vectors in the codebook, as shown in Eq. (13). Here, N represents the number of vectors (or nodes) in the latent representation Z produced by the encoder. $c_{z_i}$ is the vector of code words, corresponding to the nearest neighbors identified $z_i$ during the quantization process, and $\eta$ is the weight coefficient, typically set to 0.25, balances the loss terms.

$$\mathcal{L}_{vq} = \frac{1}{N} \sum_{z_i \in Z}^N \|\text{sg}[\![z_i]\!] - c_{z_i}\|_2^2 + \frac{\eta}{N} \sum_{z_i \in Z}^N \|\text{sg}[\![c_{z_i}]\!] - z_i\|_2^2 \tag{13}$$

The approximate counterfactual candidate generation refers to the process of using the trained VQ-CFX model to produce multiple approximate counterfactual examples for each sample in the target input set $\mathcal{D}^{Y_t}$, serving as instance-level counterfactual explanation candidates. For each sample, the factual encoder and counterfactual encoder of the VQ-CFX model generate potential representations of the input samples, which are then vector-quantized to obtain the corresponding quantized representations and codevector indices. The counterfactual quantized representations are fed into the MPNN decoder, and multiple approximate counterfactual example candidates are generated through multiple adjacency matrix samplings. Additionally, the differences between the codevector indices corresponding to the factual and counterfactual representations are compared and aggregated globally to obtain the frequency distribution of all codewords in the codebook used by the decoder to generate counterfactual examples.

### 3.3 Global counterfactual summarization algorithm

In this paper, we propose the Global Counterfactual Summarization (GCFS) algorithm to identify an optimal set of counterfactual explanation graphs from a candidate pool of approximate counterfactuals for a given target. This approach provides comprehensive global insights into how the model modifies the target prediction. The objective of model-level counterfactual explanation is to find the minimum number of counterfactual examples while maximizing the number of input samples they can interpret. This is an NP-hard problem, making it challenging to find an optimal solution in polynomial time. Inspired by the literature [30], we introduce a global counterfactual summarization algorithm that extracts model-level counterfactual explanations from a set of candidate proximity counterfactual examples, aiming to maximize global coverage. Specifically, the global counterfactual set is initially empty $\mathcal{S} \leftarrow \emptyset$. A scoring function evaluates and ranks all approximate counterfactual candidates in the target sample set, down-ranking less relevant ones. The algorithm then iteratively selects the best global counterfactuals based on a gain function, which adds counterfactuals at each iteration according to the candidate's maximal coverage gain to $\mathcal{S}$. This process continues until a specified number of optimal global counterfactuals are obtained, serving as a model-level explanation summary, as shown in Algorithm 1.

---

**Algorithm 1: Global CounterFactual Summarization**

**Input:** Candidates $\Psi$; Target size $K$; Covering function $C(\cdot)$
**Output:** Summarization $\mathcal{S}$

1. $\mathcal{R} \leftarrow \text{Sorted}(\Psi)$
2. $\mathcal{S} \leftarrow \{\}$
3. $\mathcal{S} \leftarrow \{\mathcal{R}^1\} \cup \mathcal{S}$
4. For $i$ from 2 to $\min(K, |\mathcal{R}|)$ do
5. $u \leftarrow \arg\max_{\mathcal{R}^k \in} \left( C(\mathcal{S} \cup \{\mathcal{R}^k\}) - C(\mathcal{S}) \right)$
6. $\mathcal{S} \leftarrow \{u\} \cup \mathcal{S}$
7. Endfor

8. Return $\mathcal{S}$

The scoring function evaluates the overall importance of each candidate by calculating a weighted sum of three metrics: the candidate's counterfactual Validity, Coverage, and Expressibility, as shown in Eq. (14). The validity $V(\cdot)$ and coverage functions $C(\cdot)$ are described in detail in Section 4.2. The expressibility function $E(\cdot)$ measures the proportion of high-frequency codevectors present in the quantized latent representations of the counterfactual examples generated by VQ-CFX. It is calculated as shown in Eq. (15). This function $(\cdot)$ counts the frequency of the corresponding codevectors, considered as "perturbations" in the process of generating approximate counterfactual examples. The parameter $\zeta$ is an adjustable threshold.

$$F_{score}(\mathcal{G}) = \alpha V(\mathcal{G}) + \beta C(\mathcal{G}) + (1-\alpha)E(\mathcal{G}) \tag{14}$$

$$E(\mathcal{G}) = \frac{|\{z \in Z_c^* | \text{ freq}(z) \geq \zeta\}|}{|Z_c^*|} \#(2\text{-}18)$$

The gain function calculates the contribution of the sample to the coverage of the global counterfactual set $\mathcal{S}$, as shown in Eq. (15).

$$F_{gain}(\mathcal{G}) = C(\mathcal{S} \cup \{\mathcal{G}\}) - C(\mathcal{S}) \tag{15}$$

## 4. Experiments

### 4.1 Datasets

In this paper, we train deep graph classification models and conduct global counterfactual explanation experiments using one synthetic dataset and four real-world datasets. The datasets include P5Motif, Mutagenicity [48], AIDS [49], and BBBP [50]. The statistical information for these four datasets is presented in Table 1.

Table 1 Graph dataset statistics

| Dataset | Labels | Graphs | Avg.node | Avg.edge | Features |
|---|---|---|---|---|---|
| P5Motif | 2 | 12000 | 22.45 | 26.63 | 5 |
| Mutagenicity | 2 | 4337 | 30.32 | 30.77 | 14 |
| AIDS | 2 | 2000 | 15.69 | 16.2 | 4 |
| BBBP | 2 | 2050 | 23.9 | 51.6 | 9 |

P5Motif is an artificial graph classification dataset with ground-truth explanations, which we constructed in this work. The dataset defines five node types, two graph labels, and multiple basic graph patterns associated with these labels, also known as network motifs—simple building blocks of specially designed complex networks [51]. Following existing studies [29, 52], each graph sample in P5Motif consists of a random scale-free network connected to a motif corresponding to a label. The scale-free network is a randomly generated undirected acyclic graph based on the Barabási-Albert model, with the number of nodes ranging from 5 to 20 and node types randomly assigned from five candidate types. As shown in Figure 4, the motifs for the two graph labels are designed according to the following rules: (i) Class One contains six graph patterns that satisfy mutual exclusivity, meaning that no two graph patterns can form an equivalence or containment relationship through graph matching. Each of the six graph patterns is a specific graph structure consisting of five nodes, including Motifs 1, 2, and 3, which consist of nodes of the same type, and Motifs 4, 5, and 6, which consist of nodes of multiple types. (ii) Class Two is the negative class of Class One and contains any five-node pattern other than those defined by the Class One motifs. These are obtained in practice by randomly perturbing the Class One motifs until rule (i) is violated—for example, by adding or deleting edges to change the structure or modifying node types to alter node compositions.

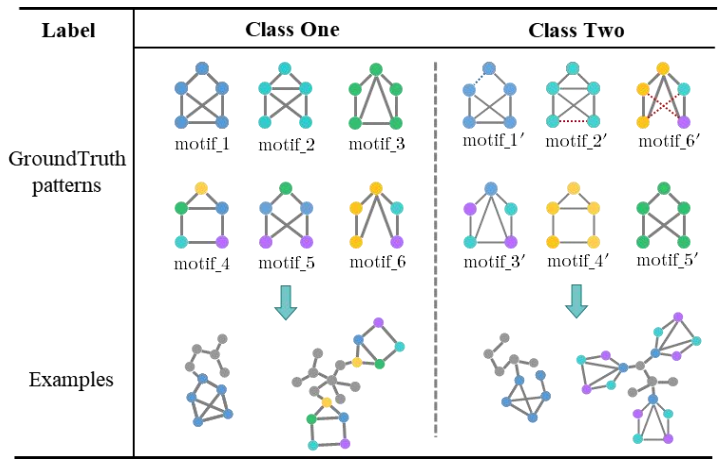

Figure 4    Motifs and examples of labels for the P5Motif dataset

Mutagenicity[48,49,53] is a molecular graph dataset that characterizes the mutagenic properties of compounds, containing two graph labels: mutagenic and non-mutagenic. Mutagenicity reflects a compound molecule's ability to induce mutations in DNA and is one of the many undesirable properties that prevent a compound from becoming a marketable drug. The dataset represents chemical-organic molecules as graphs, where each atom serves as a node, chemical bonds between atoms serve as edges, and atom types and bond types serve as node and edge attributes, respectively. Mutagenicity contains 4,337 real-world aromatic or hydrocarbon compounds, including 2,401 mutagenic and 1,936 non-mutagenic compounds composed of 14 different atom types and 4 types of chemical bonds.

The AIDS[49] dataset is a graph dataset consisting of molecular compounds, containing 2,000 samples from the Antiviral Screen Database of Active Compounds. It has two graph labels－active or inactive－indicating whether a molecule exhibits anti-HIV activity. By representing atoms as nodes and chemical bonds as edges, the AIDS dataset includes 1,600 graphs of inactive molecules and 400 graphs of active molecules.

The Blood-Brain Barrier Penetration (BBBP)[50] dataset includes 2,050 compounds related to the permeability of the blood-brain barrier, a membrane that separates circulating blood from the brain's extracellular fluid and blocks most drugs, hormones, and neurotransmitters. This property is particularly important for central nervous system drug development. The BBBP dataset uses permeability as a graph label, with atoms as nodes, chemical bonds as edges, and each node characterized by a 9-dimensional feature vector.

**4.2 Evaluation metrics**

The goal of model-level counterfactual explanation is to seek out a limited set of approximate counterfactuas that can explain the global behavior of a black-box model. The lack of graph datasets labeled with counterfactual explanations precludes the adoption of evaluation metrics for model-level factual explanations. In order to be able to quantitatively evaluate the generated model-level counterfactual explanations, this paper refers to the settings of Kosan et al[30] and uses the following global counterfactual evaluation metrics: counterfactual validity, coverage, and cost. Counterfactual validity measures the extent to which the counterfactual examples make the model predictions change. Assuming that the total number of labels in the dataset is $L$, the target label is $Y_t$, the set of model-level counterfactuals generated by the explanation method is $\Omega$, and the black-box model's normalized category probability estimate of the $k$-th counterfactual example's output is $p^k \in \mathbb{R}^L$, the counterfactual validity can be computed by using Eq. (16). Where $p_i^k$ denotes the model's prediction of the probability that the $k$-thexample belongs to the $i$-th label category, $1 \leq i, t \leq L$ and $i \neq t$. The function $\Theta(\cdot)$ crops the input values to ensure that the outputs do not exceed a certain threshold, which is uniformly set to 1.5 in this experiment.

$$Validity(\Omega) = \frac{1}{|\Omega|} \sum_k \Theta\left(\frac{p_t^k}{\max_i p_i^k}\right) \tag{16}$$

The counterfactual coverage is a core metric in the assessment of model-level counterfactual explanations, which reflects whether a set of approximate counterfactual sets $\Omega$ used to explain the global behavior of the model are representative and generic. In this paper, we consider model-level counterfactual explanations for models predicting the behavior of arbitrary target labels, and for this reason coverage is used to assess the ability of an explanation method to explain the full sample of target labels in a dataset through a limited number of counterfactual examples. For the target label $Y_t$ and the target sample set $\mathcal{D}^{Y_t}$, given a cost threshold $\delta$, the counterfactual coverage is computed as shown in Eq. (17). Where the function $\omega(\cdot)$ measures the cost of counterfactual perturbation for a individual counterfactual example to explain the target sample, this is evaluated by calculating the Graph Edit Distance (GED) between the target sample and the counterfactual. Since computing the graph edit distance accurately is an NP-hard problem, this paper employs a state-of-the-art neural approximation algorithm[54] to estimate the GED for efficient computation.

$$Coverage_\delta(\Omega, \mathcal{D}^{Y_t}) = \frac{\left|\left\{\mathcal{G} \in \mathcal{D}^{Y_t} \middle| \min_{\mathcal{G}^* \in \Omega} \omega(\mathcal{G}, \mathcal{G}^*) \leq \delta\right\}\right|}{|\mathcal{D}^{Y_t}|} \tag{17}$$

The counterfactual cost describes the minimum cost of applying a finite set of counterfactual examples to explain the global behavior of the model, which is defined as the aggregate cost of the approximate counterfactual set $\Omega$ to cover the entire set of target inputs $\mathcal{D}^{Y_t}$ as shown in Eq. (18). Where the distance function $\omega(\cdot)$ computes the minimum cost for each target sample to be explained by the counterfactual examples in $\Omega$, and the aggregation function $Aggr(\cdot)$ denotes the mean or median operation on the inputs.

$$Cost(\Omega, \mathcal{D}^{Y_t}) = Aggr\left(\left\{\min_{\mathcal{G}^* \in \Omega} \omega(\mathcal{G}, \mathcal{G}^*) \middle| \mathcal{G} \in \mathcal{D}^{Y_t}\right\}\right) \tag{18}$$

**4.3 Benchmarks and settings**

With the synthetic and real graph datasets mentioned above, we construct and train graph classification models based on graph isomorphism networks (GINs). Specifically, four GIN layers are first stacked for learning and updating graph node representations, which employ a two-layer fully connected network as a one-shot function for message aggregation with a hidden layer dimension of 64. Then a Global Attention Pooling (GAP) layer is employed to extract graph embedding representations for each sample. Finally, the normalized category probability estimates are output by a three-layer fully connected network and Softmax activation function, with each layer outputting dimensions of 64, 16, and $L$, respectively, where $L$ denotes the number of graph labels in the dataset. The training is done using five-fold cross-validation with a binary cross-entropy loss as the objective function, with a total of 300 rounds of iterations, and the predictions of the consensus model are used as the next step in interpreting the behavior of the black-box model. The classification performance of the trained deep graph model is evaluated as shown in Table 2.

Table 2  Evaluation results of classification performance (%, mean±std)

| Datasets | Accuracy | ROCAUC | F1-Score |
| --- | --- | --- | --- |
| P5Motif | 99.61±0.24 | 99.61±0.23 | 99.98±0.02 |
| Mutagenicity | 83.68±1.54 | 86.14±1.40 | 88.96±1.09 |
| AIDS | 99.51±0.56 | 98.75±1.43 | 99.72±0.37 |
| BBBP | 95.71±0.79 | 96.38±1.33 | 96.57±0.62 |

In the counterfactual explaining experiments, two current SOTA counterfactual explanatory methods, GCFExplainer[30] and CLEAR[55], are used in this paper for comparative evaluation. Among them, GCFExplainer is the only current global counterfactual explanatory method, which designs a heuristic counterfactual search process and a greedy selection algorithm to find a set of optimal model-level explanations from the global counterfactual space near the target sample set. CLEAR is a generative instance-level counterfactual explanatory method that designs counterfactual explainers based on variational graphical self-encoders capable of generating robust local counterfactual examples for arbitrary input samples. Since CLEAR is not a model-level explanatory method, it is combined with the greedy selection algorithm of GCFExplainer as CLEAR+, i.e., the local counterfactual candidates generated by CLEAR on the target set of samples are firstly applied, and the greedy selection algorithm is run according to the principle of maximal coverage gain until a finite set of a finite number of global counterfactual explanatory graphs is filtered out. The architecture and arguments of the benchmark models for the comparison were set according to those in the official code implementation, and the size of model-level counterfactual explanations that was eventually generated on each dataset was kept consistent with GCFX.

For the GCFX model proposed in this paper, its global counterfactual explanation procedure is divided into two phases: the training phase of the VQ-CFX model and the counterfactual explanation generation phase. Firstly, the VQ-CFX model is constructed with the output dimension $D_{LE}$ of 16 for the label encoder, the number of Block layers of both graph encoders is 4, and the output dimension is 32; the number of codevectors in the codebook, $N_{vq}$, is 1024, and the dimensionality of the compressed codeword vectors, $D_{vq}$, is 16; the codebook is homogeneously initialized using KMeans, with the orthogonal regularization loss weight, $w_{or}$ is 5. For the MPNN decoder, the noise distribution $\epsilon$ is the standard Gaussian distribution $N(\mu = 0, \sigma = 1)$, the decoder contains the number of decoding unit layers $T$ is 4, the input and output edge reconstruction representations $E^i$ have the feature dimensions of [2,8,16,8,4], and the node reconstruction representations $H^i$ have the feature dimensions of [48,32,32,32,16], and $0 \leq i \leq T$. The two output dimensions of the predictor modules AMP and NFP correspond to the adjacency matrix and node feature dimensions of the original samples of the dataset, respectively. The VQ-CFX model is trained using the AdamW optimizer with a learning rate of lr of 0.001 and is scheduled using OneCycleLR, with a preheating factor of 0.1 and a decay factor of 10.In addition, the number of iterations epoch is set as 400 and the data batch size batch_size is 128.

In the counterfactual explanation generation phase, the proposed GCFX method first applies the trained VQ-CFX to generate approximate counterfactual examples for each sample in the target input set, and then runs the GCFS algorithm to extract the global counterfactual summarization from the counterfactual candidate set, which serves as the final model-level explanation. The capacity K of the global counterfactual summarization is 50 for real datasets and 20 for synthetic datasets.

### 4.4 Results and analysis
**Close counterfactual example generation**

An example of the approximate counterfactual generated for the target sample set by applying the above explanatory method on the synthetic dataset P5Motif with the target label $Y_t = 0$ and the counterfactual label $Y_c = 1$ is shown in Figure 5. Where the probability prob denotes the probability estimate that the corresponding example makes the model predict the counterfactual label, the metric cost evaluates the counterfactual perturbation cost between the corresponding example and its input samples, and identifies the types of perturbations such as addition, deletion, and modification targeting the nodes or edges by red, green, and orange colors, respectively.

According to the visualization results, GCFX is able to generate similar counterfactual examples for both simple and complex input samples, with a low counterfactual perturbation cost while allowing the model to predict counterfactual labels with high probability. As seen, GCFX reconstructs similar input samples as much as possible to ensure minimal perturbation of the counterfactuals, and is able to capture changes in key structures (e.g., Motifs) in the samples to effectively change the model predictions. GCFExplainer, which is based on heuristic search with graph editing perturbations, is also able to generate effective counterfactuals, but the quality of the found counterfactuals depends on its full exploration of the counterfactual space and randomized wandering paths. Therefore, GCFExplainer finds approximate counterfactuas for the simple sample (1) that require only one effective graph edit, while for sample (2) it spends more costly counterfactual explorations, and in particular generates additional graph edits for the label-independent scale-free network part of the graph. Finally, the CLEAR method generates high probability counterfactual examples while incurring a non-negligible perturbation cost on the input samples. The method pushes the model to predict counterfactual labels at the cost of excessive input perturbation and prefers to generate sparse or even non-connected counterfactual examples.

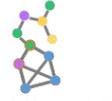

Figure 5 Results of approximate counterfactual example generation for the synthetic dataset

The approximate counterfactual explanation results of the deep graph model on three real molecular datasets with domain knowledge constraints are shown in Figure 6. In particular, the GCFExplainer method has only a limited number of editing operations that can produce legitimate molecular structures during counterfactual exploration, while most of the random edits result in atomic valence violations in the generated counterfactual structures, such as the difference between the "tetravalent" fluorine atom (F) in the first row and the "hexavalent" carbon atom (C) in the second row. CLEAR behaves similarly in that it is able to generate explanations with high counterfactual probabilities but cannot guarantee the reasonability of the counterfactual molecular structures, and the property of always tending to generate sparse neighbor matrices also leads to more nonsensical molecular fragments in its counterfactual examples. In contrast, the GCFX proposed in this paper generates legitimate and effective counterfactual examples for multiple input samples of each dataset, which also shows that the VQ-CFX model it adopts can well capture the local structural patterns in the real data distributions, thus ensuring that the generated counterfactuals satisfy the validity and proximity while taking into account the legitimacy of the domain constraints of the real data.

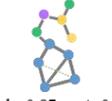

Figure 6 Results of approximate counterfactual example generation for molecular datasets

**Model-level counterfactual explanations**

Table 3 demonstrates the evaluation results of the model-level counterfactual explanations of GCFX with the benchmark methods on different datasets, and the samples predicted to be labeled 0 in each dataset are taken to construct the target input set, respectively. According to the results, it can be analyzed that the GCFX method achieves the best performance in both counterfactual validity and coverage metrics, and meets with a low perturbation cost on the overall target sample set. In addition, GCFX based on depth graph generation is able to generate approximate counterfactual candidate sets efficiently with linear time complexity, and the model-level explanations generated by GCFX based on quantified potential distributions of counterfactuals are easier to generalize to new unknown samples than the heuristic methods that rely on the target sample set to explore the counterfactuals.

Tabel 3 Evaluation results for model-level counterfactual explanations

| Datasets | Methods | Validity | Coverage(%) | Cost |
|---|---|---|---|---|
| P5Motif | GCFExplainer | 1.08 | 33.19 | 0.13 |
| | CLEAR+ | 1.15 | 20.41 | 0.23 |
| | GCFX | 1.22 | 41.26 | 0.11 |
| Mutagenicity | GCFExplainer | 1.12 | 53.27 | 0.15 |
| | CLEAR+ | 1.27 | 23.74 | 0.25 |
| | GCFX | 1.31 | 59.83 | 0.19 |
| AIDS | GCFExplainer | 1.05 | 27.28 | 0.17 |
| | CLEAR+ | 1.09 | 11.43 | 0.28 |
| | GCFX | 1.15 | 31.94 | 0.21 |
| BBBP | GCFExplainer | 1.10 | 28.79 | 0.16 |
| | CLEAR+ | 1.14 | 13.79 | 0.26 |
| | GCFX | 1.17 | 29.87 | 0.18 |

## 5. Conclusions

To address model-level counterfactual explanations for deep graph classification models, this paper proposes a generative global counterfactual explainer, GCFX, designed to elucidate the global predictive behavior of black-box models from a global counterfactual perspective. By combining deep graph generation with vector quantization techniques, GCFX effectively learns latent distributions of real data and generates high-quality approximate counterfactual examples. It then identifies a set of optimal model-level counterfactual explanatory graphs through global counterfactual summarization, further revealing the global decision logic underlying the model's target behavior. Experimental validation on both synthetic and real datasets demonstrates that the model-level counterfactual explanations produced by GCFX exhibit superior performance in terms of global validity and coverage, while maintaining a low explanation cost. Compared to existing generative counterfactual methods, GCFX generates approximate counterfactual examples that are closer to the input samples. Furthermore, relative to existing heuristic global explanatory methods, GCFX efficiently produces counterfactual explanations that adhere to data domain constraints with minimal black-box model queries. These results demonstrate the feasibility of integrating deep graph generation with vector quantization techniques for model-level counterfactual explanation applications, as well as the potential of global counterfactual summarization to enhance the transparency and reliability of classification decisions in deep graph models.